\documentclass[journal]{IEEEtran}
\usepackage{cite}
\usepackage{subfig}
\usepackage{amsmath,amssymb,amsfonts,amsthm}
\usepackage{algorithmic}
\usepackage{graphicx}
\usepackage{hyperref}
\usepackage{multirow}
\usepackage{multicol}
\usepackage{booktabs}
\hypersetup{colorlinks=true,urlcolor=red}
\usepackage{bm} 
\usepackage{xspace}

\newcommand{\xb}{{\blmath x}}
\newcommand{\yb}{{\blmath y}}

\newcommand{\Hc}{\mathcal{H}}

\newcommand{\Xc}{\mathcal{X}}
\newcommand{\Yc}{\mathcal{Y}}

\newcommand{\Rd}{{\mathbb R}}

\newcommand{\Kd}{\mathbb{K}}

\newcommand{\beq}{\begin{equation}}
\newcommand{\eeq}{\end{equation}}
\newcommand{\beqa}{\begin{eqnarray}}
\newcommand{\eeqa}{\end{eqnarray}}
\newcommand{\Fc}{{\mathcal F}}

\renewcommand{\xb}{{x}}
\renewcommand{\yb}{{y}}

\begin{document}

\title{CycleQSM: Unsupervised QSM Deep Learning using Physics-Informed CycleGAN}
\date{\vspace{-4ex}}

\author{Gyutaek Oh, Hyokyoung Bae, Hyun-Seo Ahn, Sung-Hong Park, and~Jong~Chul~Ye,~\IEEEmembership{Fellow,~IEEE}
\thanks{G. Oh, H. Bae, H-S. Ahn, S-H. Park, and J. C. Ye are with the Department of Bio and Brain Engineering, 
		Korea Advanced Institute of Science and Technology (KAIST), 
		Daejeon 34141, Republic of Korea (e-mail: \{okt0711, ippuny3, hyunsea, sunghongpark, jong.ye\}@kaist.ac.kr). 
		J.C. Ye is also with the Department of Mathematical Sciences, KAIST.} 
}

\maketitle

\begin{abstract}
Quantitative susceptibility mapping (QSM) is a useful magnetic resonance imaging (MRI) technique which provides spatial distribution of magnetic susceptibility values of tissues.
QSMs can be obtained by deconvolving the dipole kernel from  phase images, but the spectral nulls in the dipole kernel make the inversion ill-posed.
In recent times, deep learning approaches have shown a comparable QSM reconstruction performance as the classic approaches, despite the fast reconstruction time. 
Most of the existing deep learning methods are, however, based on supervised learning, so  matched pairs of input phase images and the ground-truth maps are needed.
Moreover, it was reported that the supervised learning often leads to underestimated QSM values.
To address this, here we propose a novel unsupervised QSM deep learning method  using physics-informed cycleGAN, which is derived from optimal transport perspective.
In contrast to the conventional cycleGAN, our novel cycleGAN  has only one generator and one discriminator thanks to the known dipole kernel.
Experimental results confirm that the proposed method provides more accurate QSM maps compared to the existing deep learning approaches,
and provide competitive performance to the best classical approaches despite the ultra-fast reconstruction.
\end{abstract}


\IEEEpeerreviewmaketitle

\section{Introduction}\label{sec:introduction}
\IEEEPARstart{Q}{uantitative} susceptibility mapping (QSM) is a magnetic resonance imaging (MRI) technique that measures magnetic susceptibility values from MRI phase images \cite{haacke2015quantitative,wang2015quantitative}.
QSM offers image contrast that is sensitive to iron deposition, which is useful for the diagnosis of various diseases \cite{bilgic2012mri,deistung2013quantitative,langkammer2013quantitative,barbosa2015quantifying}.


From a mathematical point of view, a phase image can be expressed by the convolution between the magnetic susceptibility distribution and the dipole-shaped kernel.
Thus, a magnetic susceptibility distribution can be obtained by deconvolution: for example, dividing the phase by the dipole kernel in the Fourier domain.
However, the dipole kernel in the Fourier domain has zeros along the conical surface, which makes the dipole inversion an ill-posed inverse problem.

To overcome this limitation, various dipole inversion approaches have been investigated.
Calculation of susceptibility through multiple orientation sampling (COSMOS) \cite{liu2009calculation} is considered as the gold standard algorithm for the dipole inversion.
COSMOS restores accurate QSM from multiple head orientation data.
Unfortunately, the acquisition of MR data  along multiple head orientations for COSMOS reconstruction takes too much subject's time and effort.

Accordingly, algorithms have been developed for dipole inversion from single head orientation data.
For example, filling the area around the conical surface with a threshold value is one of the method for dipole inversion \cite{shmueli2009magnetic,schweser2013toward}.
Also, the regularization by edge information of the magnitude image can mitigate the ill-posedness of dipole inversion \cite{liu2011morphology,liu2012accuracy}.
In addition, dipole inversion algorithms based on compressed sensing \cite{wu2012whole,ahn2020quantitative} have been studied.
However, the streaking artifact in the reconstructed QSM and the difficulty of hyper-parameter tuning are limitations of the above algorithms.

Recently, inspired by the successes of deep learning approaches for medical image reconstruction \cite{kang2017deep,gong2018deep,cha2020geometric,khan2020adaptive}, deep learning algorithms have also been extensively studied for QSM reconstruction \cite{yoon2018quantitative,bollmann2019deepqsm,chen2020qsmgan,gao2020xqsm,liu2020deep,liu2020weakly,polak2020nonlinear}.
These algorithms show comparable performance as the classical methods despite the fast computational time.
However, most of the existing deep learning methods for QSM reconstruction are based on supervised learning which requires matched pairs of phase images and ground-truth QSM labels.
Nonetheless, it has been reported that the reconstructed QSM values are often underestimated \cite{jung2020overview}.

To address this, here we propose a novel physics-informed cycleGAN approach for unsupervised QSM reconstruction, inspired by our recent theory of optimal transport driven cycleGAN (OT-cycleGAN) \cite{sim2019optimal,lim2020cyclegan,oh2020unpaired}.
Since the forward mapping from the susceptibility map to the phase image is given as a convolution with the known dipole kernel, our mathematical theory \cite{sim2019optimal} leads to a simple OT-cycleGAN architecture composed of a single pair of generator and discriminator.
The resulting architecture, dubbed {\em CycleQSM}, exhibits faster and more stable training than the original cycleGAN composed of two sets of deep learning based generator and discriminators.
Furthermore, it provides better performance than the existing deep learning algorithms, and results in the comparable performance to the best available classical approach.

This paper is organized as follows.
Section \ref{sec:related works} reviews the classical  and deep learning approaches  for QSM reconstruction.
Section \ref{sec:theory} then describes the physics for dipole inversion and proposes physics-informed OT-cycleGAN for QSM reconstruction.
Next, experimental data sets and details of our method are presented in Section \ref{sec:method}.
Our experimental results and the discussion are shown in Section \ref{sec:result} and Section \ref{sec:discussion}, respectively.
Finally, Section \ref{sec:conclusion} concludes our work.

\section{Related Works}\label{sec:related works}
\subsection{Classical Methods for QSM Reconstruction}
Various methods have been developed to solve the ill-posed dipole inversion problem.
For example, thresholded $k$-space division (TKD) \cite{shmueli2009magnetic,schweser2013toward} is a method of filling values near the conical surface of the dipole kernel with a threshold value.
Instead of filling the dipole kernel with a specific value, some methods used additional information for the dipole inversion.
Schweser et al. \cite{schweser2012quantitative} proposed homogeneity enabled incremental dipole inversion (HEIDI) using homogeneity information from gradient echo phase images.
Morphology enabled dipole inversion (MEDI) \cite{liu2011morphology,liu2012accuracy} is a method that exploits the structural consistency between the magnitude image and the reconstructed quantitative susceptibility map.
Li et al. \cite{li2015method} introduced a method based on a sparse linear equation and least squares algorithm (iLSQR) to remove streaking artifacts in reconstructed QSM.
Compressed sensing \cite{CaRoTa06,donoho2006compressed} based algorithms were also examined.
Annihilating filter-based low-rank Hankel matrix approach (ALOHA) for QSM \cite{ahn2020quantitative} exploits the sparsity of the data in a certain transform domain to interpolate the missing $k$-space data in the dipole spectral null.

Although above algorithms showed high performance, there are limitations of the above algorithms, such as streaking artifacts, difficulties in optimizing hyper-parameters, and high computational complexity.

\subsection{Deep Learning Methods for QSM Reconstruction}
To overcome the limitations of traditional methods, QSM reconstruction algorithms based on deep learning have been investigated.
Yoon et al. \cite{yoon2018quantitative} and Bollman et al. \cite{bollmann2019deepqsm} proposed QSMnet and deepQSM, respectively, which are 3D U-net \cite{ronneberger2015u} structures designed for QSM reconstruction.
By using of COSMOS images as ground-truth QSM labels, they showed comparable results to those of classical approaches.
Chen et al. \cite{chen2020qsmgan} suggested QSMGAN where the adversarial loss was utilized, and Gao et al. \cite{gao2020xqsm} introduced xQSM where the octave convolutional layers were applied.
Polak et al. \cite{polak2020nonlinear} also introduced nonlinear dipole inversion with deep learning using a variational neural network, which combines optimization of nonlinear QSM data model and the data fidelity term.

Although these methods have shown improved image quality compared to conventional methods, the requirement of matched pairs containing phase images and susceptibility maps is a limitation of aforementioned supervised methods.
Moreover, in a recent review paper \cite{jung2020overview}, the authors pointed out that one of the biggest issues of supervised learning approach is the generalization error that happens when test data have different characteristics to training data.
For example, compared with the conventional QSM, deep learning QSM results underestimated the susceptibility values when the susceptibility range in the training data differs from the test data.
Moreover, this effect is shown severe when training with synthetic data, which lacks enough variability and structure \cite{jung2020overview}.

Recently, some deep learning algorithms based on weakly supervised or unsupervised learning were introduced.
Liu et al. \cite{liu2020weakly} proposed weakly supervised learning for QSM reconstruction (wTFI).
wTFI reconstructs QSM without background field removal, and enables the restoration of susceptibility values near the edges of the brain, which can disappear with background field removal.
Liu et al. \cite{liu2020deep} proposed unsupervised QSM reconstruction method (uQSM) by taking advantage of nonlinear data consistency loss and the total variation loss.

\subsection{Contributions}
The contributions of our work are as follows.
\begin{itemize}
\item Unlike the supervised learning approaches \cite{yoon2018quantitative,bollmann2019deepqsm,chen2020qsmgan,gao2020xqsm,polak2020nonlinear}, our method does not require matched QSM labels for the training so that the trained model is less sensitive to the lack of training data with enough variability and structures.
Accordingly, the underestimation issue of QSM values in the supervised learning can be largely overcome.
\item In contrast to the existing unsupervised learning approaches \cite{liu2020weakly,liu2020deep}, our method learns statistical properties of unpaired QSM labels during training, which makes the training more stable.
This leads to the more accurate QSM map estimation with less outliers.
\item Unlike the classical approaches \cite{shmueli2009magnetic,schweser2013toward,liu2011morphology,liu2012accuracy,wu2012whole,ahn2020quantitative} and deep image prior (DIP)\cite{ulyanov2018deep}, our method is a feed-forward neural network that provides instantaneous reconstruction once the cycleGAN training is done.
This makes the algorithm very practical.
\end{itemize}

\begin{figure*}[!t]
    \centerline{\includegraphics[width=0.9\linewidth]{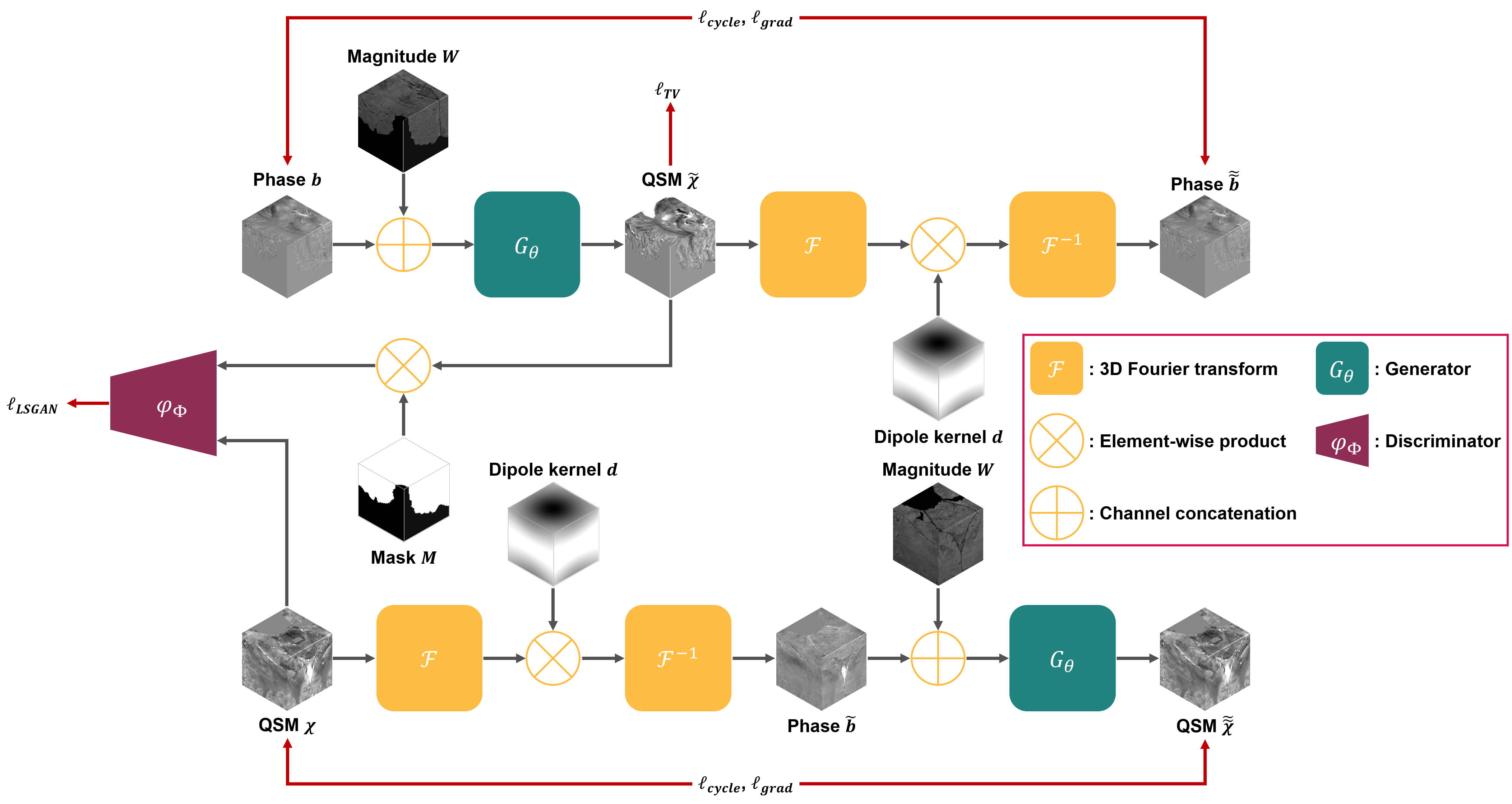}}
    \caption{CycleQSM architecture.
    Our cycleGAN architecture has only one generator and one discriminator because the generator for producing phase images is replaced by the element-wise product with the known dipole kernel in the Fourier domain.
    Here, generated susceptibility map is multiplied by the brain mask before entering the discriminator.}
    \vspace{-0.5cm}
    \label{fig:QSM_cycleGAN}
\end{figure*}


\section{Theory}\label{sec:theory}
\subsection{Dipole Inversion}
When the tissue is brought into the magnetic field, the tissue becomes magnetized.
Magnetic susceptibility $\chi$ is the quantitative measure of the degree of magnetization.
In MRI, the magnetization of tissues generates the magnetic perturbation along the main magnetic field.
This magnetic perturbation, or the phase signal can be represented as follows:
\begin{eqnarray}\label{eq:fwd_fourier}
b(\Vec{r})=d(\Vec{r})*\chi(\Vec{r}),\quad r\in \Rd^3
\end{eqnarray}
where $b$ is the phase signal, $d$ is the dipole kernel, and $*$ is the convolution operation.
Here, the dipole kernel $d$ is represented by
$$d(\Vec{r})=\frac{1}{4\pi}\frac{3\cos^2\theta-1}{|\Vec{r}|^3}$$
where $\theta$ is the angle between $\Vec{r}$ and the main magnetic field, whose Fourier domain spectrum is given by
\begin{eqnarray}\label{eq:dipole}
\hat d(\Vec{k})=\frac{1}{3}-\frac{k^2_z}{|\Vec{k}|^2}
\end{eqnarray}
where $\Vec{k}=[k_x,k_y,k_z]^\intercal$ is a $k$-space vector.

Accordingly, one of the simplest dipole inversions could be achieved by element-wise division in the Fourier domain.
\begin{eqnarray}\label{eq:dipole_inversion}
\hat\chi(\Vec{k})=\frac{\hat b(\Vec{k})}{\hat d(\Vec{k})}
\end{eqnarray}
However, \eqref{eq:dipole_inversion} is not stable because $\hat D(\Vec{k})$ has zero values at the conical surface ($\Vec{k}^2=3k^2_z$).

\subsection{Optimal Transport Driven CycleGAN}
In our recent mathematical theory of optimal transport driven cycleGAN (OT-cycleGAN) \cite{sim2019optimal,lim2020cyclegan,oh2020unpaired}, we revealed that various forms of cycleGAN architecture can be obtained from the dual formulation of an optimal transport problem, in which the transport cost is the sum of the distances in the measurement and image domains.
In particular, if the forward mapping is known from the imaging physics as in QSM, the resulting OT-cycleGAN architecture can be significantly simplified.
Here, we briefly review the theory to apply it for the QSM reconstruction.
 
Consider the following measurement model:
\begin{eqnarray}
\yb&=&\Hc \xb \ , 
\end{eqnarray}
where $\yb \in \Yc$ and $\xb \in \Xc$ denote the measurement and the unknown image, respectively, and $\Hc : \Xc \mapsto \Yc$ is the {\em known} deterministic imaging operator. 
In contrast to the supervised learning where the goal is to learn the relationship between the image $\xb$ and measurement $\yb$ pairs, in the unsupervised learning framework there are no matched image-measurement pairs. 
Still we could have sets of images and unpaired measurements, so our goal is to match the probability distributions rather than each individual samples. 
This can be done by finding transportation maps that transport the probability measures between the two spaces.

Specifically, suppose that the target image space $\Xc$ is equipped with a probability measure $\mu$, whereas the measurement space $\Yc$ is with a probability measure $\nu$.
Then, 
we can see that the mass transport from $(\Xc,\mu)$ to $(\Yc,\nu)$ is performed by the forward operator $\Hc$, and the mass transportation from the measurement space $(\Yc,\nu)$ to the image space $(\Xc,\mu)$ is done by a generator $G_\Theta: \Yc \mapsto \Xc$, parameterized by $\Theta$.
Then, the following transportation cost is proposed for the optimal transport problem \cite{sim2019optimal}:
\begin{eqnarray}\label{eq:tcost}
c(\xb,\yb;\Theta):=\|\yb-\Hc\xb\|+\|G_\Theta(\yb)-\xb\|
\end{eqnarray}
which denotes the sum of the distance between a training sample and a transported sample in each space. 
Rather than minimizing the sample-wise cost using \eqref{eq:tcost}, the goal of the optimal transport is to minimize the {\em average} transport cost.
More specifically, the optimal transport problem is formulated to find the joint distribution $\pi$ that leads to the minimum average transport cost:
\begin{align}\label{eq:unsupervised}
\inf\limits_{\pi \in \Pi(\mu,\nu)}\int_{\Xc\times \Yc}c(\xb,\yb;\Theta) d\pi(\xb,\yb) 
\end{align}
where $\Pi(\mu,\nu)$ is the set of joint measures whose marginal distributions in $\Xc$ and $\Yc$ are $\mu$ and $\nu$, respectively.

In \cite{sim2019optimal,lim2020cyclegan,oh2020unpaired}, the geometric meaning of the optimal transport using \eqref{eq:tcost} is explained.
More specifically, if the first term in \eqref{eq:tcost} is only used, the optimal transport is to find the joint probability that minimizes the distance in the empirical distribution $\nu$ and the so-called ``push-forward measure'' $\nu_\Hc$.
Similarly, if the second term in \eqref{eq:tcost} is only used, the OT problem is to minimizes the distance in the empirical distribution $\mu$ and the push-forward measure $\mu_G$.
By using both terms in \eqref{eq:tcost}, our OT formulation finds the joint measure that minimizes the sum of the two distances in the measurement and the image spaces.

Using the transportation cost in \eqref{eq:tcost}, 
the Kantorovich dual formulation \cite{villani2008optimal} is given by \cite{sim2019optimal}
\begin{eqnarray}\label{eq:minmax}
\min_{\Theta}\Kd(\Theta,\Hc)=\min_{\Theta}\max_{\Phi}\ell(\Theta;\Phi)
\end{eqnarray}
where
\begin{eqnarray}
\ell(\Theta;\Phi)=\gamma\ell_{cycle}(\Theta)+\ell_{WGAN}(\Theta;\Phi)
\end{eqnarray}
where $\gamma$ is a suitable hyper-parameter, $\ell_{cycle}$ is the cycle-consistency loss, $\ell_{WGAN}$ is the Wasserstein GAN loss \cite{martin2017wasserstein}.
More specifically, $\ell_{cycle}$ and $\ell_{WGAN}$ are given by
%
\begin{eqnarray}\label{eq:cycle_simple}
\begin{aligned}
\ell_{cycle}(\Theta)=\int_\Xc\|\xb-G_\Theta(\Hc\xb)\|d\mu(\xb)\\
+\int_\Yc\|\yb-\Hc G_\Theta(\yb)\|d\nu(\yb)\ ,
\end{aligned}
\end{eqnarray}
\begin{eqnarray}\label{eq:wgan_simple}
\begin{aligned}
&\ell_{WGAN}(\Theta;\Phi)\\
&=\int_\Xc\varphi_\Phi(\xb)d\mu(\xb)-\int_\Yc\varphi_\Phi(G_\Theta(\yb))d\nu(\yb)\ .
\end{aligned}
\end{eqnarray}
Note that there exists only a single discriminator $\varphi_\Phi$, since the forward operator $\Hc$ is assumed known, so there is no need to compete with the forward operator.
This makes the cycleGAN architecture simple, as will be explained later.

\begin{table*}[!hbt]
	\centering
	\caption{Data sets for experiments}
	\label{tbl:dataset}
	\resizebox{0.85\textwidth}{!}{
		\begin{tabular}{c | c | c  c  c  c } 
			\toprule
			\multicolumn{2}{c|}{ } 				                & 2016 QSM Challenge            & Cornell	                    & ALOHA-QSM                     & 2019 QSM Challenge            \\ \midrule\midrule
			\multirow{2}*{Resolution}	& Voxel pitch (mm$^3$)  & 1.06$\times$1.06$\times$1.06  & 0.9375$\times$0.9375$\times$1 & 0.75$\times$0.75$\times$1.1   & 0.64$\times$0.64$\times$0.64  \\
			\ 						    & Matrix size	        & 160$\times$160$\times$160	    & 256$\times$256$\times$146		& 240$\times$320$\times$128     & 164$\times$205$\times$205     \\ \midrule
			\multirow{4}*{No. of Cases}	& Total                 & 1			                    & 1			                    & 5			                    & 4	(2 SNR / 2 Contrast)        \\
			\ 						    & QSM Ground-truth		& 1 (COSMOS)			        & 1 (COSMOS)		            & 0	                            & 2 (Simulation)                \\
				                        & Training	            & 1			                    & 0			                    & 3				                & 4                             \\
			\						    & Test		            & 0			                    & 1			                    & 2				                & 0                             \\ \bottomrule
		\end{tabular}
	}
	\vspace{-0.5cm}
\end{table*}

\subsection{CycleQSM}
In this section, we will see how our OT-cycleGAN formulation can be used for QSM.
In the dipole inversion, a phase image $b\in B$ and a susceptibility map $\chi\in X$ correspond to a noisy measurement and an unobserved image, respectively.
Therefore, the forward model of the dipole inversion can be formulated as
\begin{eqnarray}\label{eq:forward}
b=\Fc^{-1}\hat d\Fc\chi
\end{eqnarray}
where $\Fc$ and $\Fc^{-1}$ are 3D Fourier transform and 3D inverse Fourier transform, respectively.
By identifying $\yb:=b,\xb:=\chi$ and $\Hc:=\Fc^{-1}\hat d\Fc$, the cycle-consistency loss \eqref{eq:cycle_simple} and the GAN loss \eqref{eq:wgan_simple} can be represented by
\begin{eqnarray}\label{eq:cycle_qsm}
\begin{aligned}
\ell_{cycle}(\Theta)=\int_X\|\chi-G_\Theta(\Fc^{-1}\hat d\Fc\chi)\|d\mu(\chi)\\
+\int_B\|b-\Fc^{-1}\hat d\Fc\ G_\Theta(b)\|d\nu(b)\ ,
\end{aligned}
\end{eqnarray}
and
\begin{eqnarray}\label{eq:wgan_qsm}
\begin{aligned}
&\ell_{WGAN}(\Theta;\Phi)\\
&=\int_X\varphi_\Phi(\chi)d\mu(\chi)-\int_B\varphi_\Phi(G_\Theta(b))d\nu(b)\ .
\end{aligned}
\end{eqnarray}

Although these costs can be used directly for the QSM reconstruction, we provide additional modifications for a better performance. 
First, we employ the least squares GAN (LSGAN) \cite{mao2017least} instead of WGAN loss for faster and more stable training.
The link between WGAN and LSGAN was also explained in \cite{lim2020cyclegan}.
Next, 
we add the gradient difference loss ($\ell_{grad}$) proposed in \cite{yoon2018quantitative} to maintain the edge information.
\begin{eqnarray}\label{eq:gradient}
\begin{aligned}
\ell_{grad}(\Theta)=\int_X\|\nabla\chi-\nabla G_\Theta(\Fc^{-1}\hat d\Fc\chi)\|d\mu(\chi)\\
+\int_B\|\nabla b-\nabla\Fc^{-1}\hat d\Fc\ G_\Theta(b)\|d\nu(b)
\end{aligned}
\end{eqnarray}
where $\nabla$ is a gradient operator for the 3D volume.
To preserve image details and remove noise in reconstructed QSM, we also employ the total variation (TV) loss ($\ell_{TV}$).
\begin{eqnarray}\label{eq:TV}
\ell_{TV}(\Theta)=\int_B\|\nabla G_\Theta(b)\|d\nu(b)\ .
\end{eqnarray}
Therefore, the final cost function for CycleQSM can be formulated as
\begin{eqnarray}\label{eq:cost_final}
\begin{aligned}
\ell(\Theta;\Phi)=&\gamma\ell_{cycle}(\Theta)+\ell_{LSGAN}(\Theta,\Phi)\\
&+ \eta\ell_{grad}(\Theta)+\rho\ell_{TV}(\Theta)
\end{aligned}
\end{eqnarray}
where $\gamma$, $\eta$, and $\rho$ are appropriate hyper-parameters.
Note that the additional losses in \eqref{eq:gradient} and \eqref{eq:TV} are average values with respect to the {\em marginal} distributions.
Therefore, the addition of these terms does not change the optimal transport interpretation.
In fact, as analyzed in \cite{cha2020unpaired}, \eqref{eq:cost_final} can be obtained as a dual formulation of the optimal transport problem with the following transportation cost:
\begin{align*}
&c_{QSM}(b,\chi;\Theta)\\
:=&~\|b-\Fc^{-1}\hat d\Fc\chi\|+\|G_\Theta(b)-\chi\| \\
&+ \eta \left(\|\nabla\chi-\nabla G_\Theta(\Fc^{-1}\hat d\Fc\chi)\|+\|\nabla b-\nabla\Fc^{-1}\hat d\Fc\ G_\Theta(b)\|\right)\\
&+\rho\|\nabla G_\Theta(b)\|
\end{align*}
where $b\in \Yc$ and $\chi \in \Xc$ with probability measures $\nu$ and $\mu$, respectively.

\section{Method}\label{sec:method}
\subsection{Experimental Data}
To train and test our algorithm, we use data sets from different sources.
First data set is in vivo human brain data that was provided for 2016 QSM challenge \cite{langkammer2018quantitative}.
This data set was acquired from a healthy volunteer by 3T Siemens scanner with 3D gradient echo (GRE) sequence.
Acquisition parameters of 2016 QSM challenge data are as follows: resolution = 1.06$\times$1.06$\times$1.06 mm$^3$, matrix size = 160$\times$160$\times$160, TR/TE = 25/35 ms.
This data set also contains ground-truth QSM which is acquired by COSMOS with 12 different orientation data.

Next, we use in vivo healthy human data from the Cornell MRI Research Lab.
This data set was collected using multi-echo GRE sequence with 3T GE system.
The acquisition parameters are as follows: resolution = 0.9375$\times$0.9375$\times$1 mm$^3$, matrix size = 256$\times$256$\times$256, TR/TE$_1$ = 55/5 ms, $\Delta$TE = 5 ms.
QSM is also obtained by COSMOS with 5 different head orientation data.

The third data set was obtained from five healthy volunteers using 3T Siemens system.
This data set was used for ALOHA-QSM \cite{ahn2020quantitative}, and acquired with the following parameters: resolution = 0.75$\times$0.75$\times$1.1 mm$^3$, matrix size = 240$\times$320$\times$128, TR/TE$_1$ = 43/9.35 ms, $\Delta$TE = 8.94 ms.

Last, we used the data for 2019 QSM challenge \cite{marques2020qsm}.
This data set was obtained through forward simulation based on in vivo human brain data acquired on 7T MR system.
The acquisition parameters for 2019 challenge data are as follows: resolution = 0.64$\times$0.64$\times$0.64 mm$^3$, matrix size = 164$\times$205$\times$205, TR/TE$_1$ = 50/4 ms, $\Delta$TE = 8 ms.
The local field data in 2019 challenge data set were simulated with two different contrast levels and two signal-to-noise ratio (SNR) levels, so total four local field maps (Sim1Snr1, Sim1Snr2, Sim2Snr1, Sim2Snr2) were provided.
In addition, ground-truth QSM data were provided for two different contrast levels.

Table \ref{tbl:dataset} shows the name and number of volumes of each data set.
The 2016 QSM challenge and 2019 QSM challenge data in Table \ref{tbl:dataset} are used to train our network as well as other deep learning approaches.
For unsupervised learning, we do not need a matched reference data.
So three local field maps of the ALOHA-QSM data set are also used for training and the remaining two local field maps are used for inference.
Furthermore, the Cornell data set is used for the quantitative evaluation of our algorithm.

For data preprocessing, brain mask extraction using the brain extraction tool (BET) \cite{smith2002fast}, multi-coil phase combination by Hermitian inner product (HiP) \cite{bernstein1994reconstructions}, and multi-echo phase correction by nonlinear frequency map estimation \cite{de2008quantitative,kressler2009nonlinear,liu2013nonlinear} are performed.
Laplacian-based phase unwrapping \cite{schofield2003fast,li2011quantitative} is then applied, and then background phase removal by sophisticated harmonic artifact reduction for phase data with varying spherical kernel (V-SHARP) \cite{wu2012whole,li2014integrated} is executed.

\subsection{Network Architectures}
Fig. \ref{fig:QSM_cycleGAN} illustrates the architecture of cycleQSM.
Since the forward mapping is described by the deterministic dipole kernel in \eqref{eq:dipole}, cycleQSM has only one pair of the generator and discriminator.
As a result, the training of our network can be more stable and faster than the conventional cycleGAN. 
To provide more information to the generator, the magnitude image is concatenated with the phase image.
The dipole kernel for each training step is generated depending on the resolution and direction of the main magnetic field of each data.
Accordingly, all the data set in Table \ref{tbl:dataset} with different spatial resolution can be fully utilized for training cycleQSM. 

In Fig. \ref{fig:QSM_cycleGAN}, the reconstructed QSM is multiplied by the brain mask before going through the discriminator.
Without the brain mask, we find that the discriminator distinguishes real and fake QSM by observing the artifact outside the brain mask region.
We find that the multiplying the brain mask can stabilize the training process and reduce the artifact outside of the brain region.

\begin{figure}[!t]
    \centerline{\includegraphics[width=0.85\linewidth]{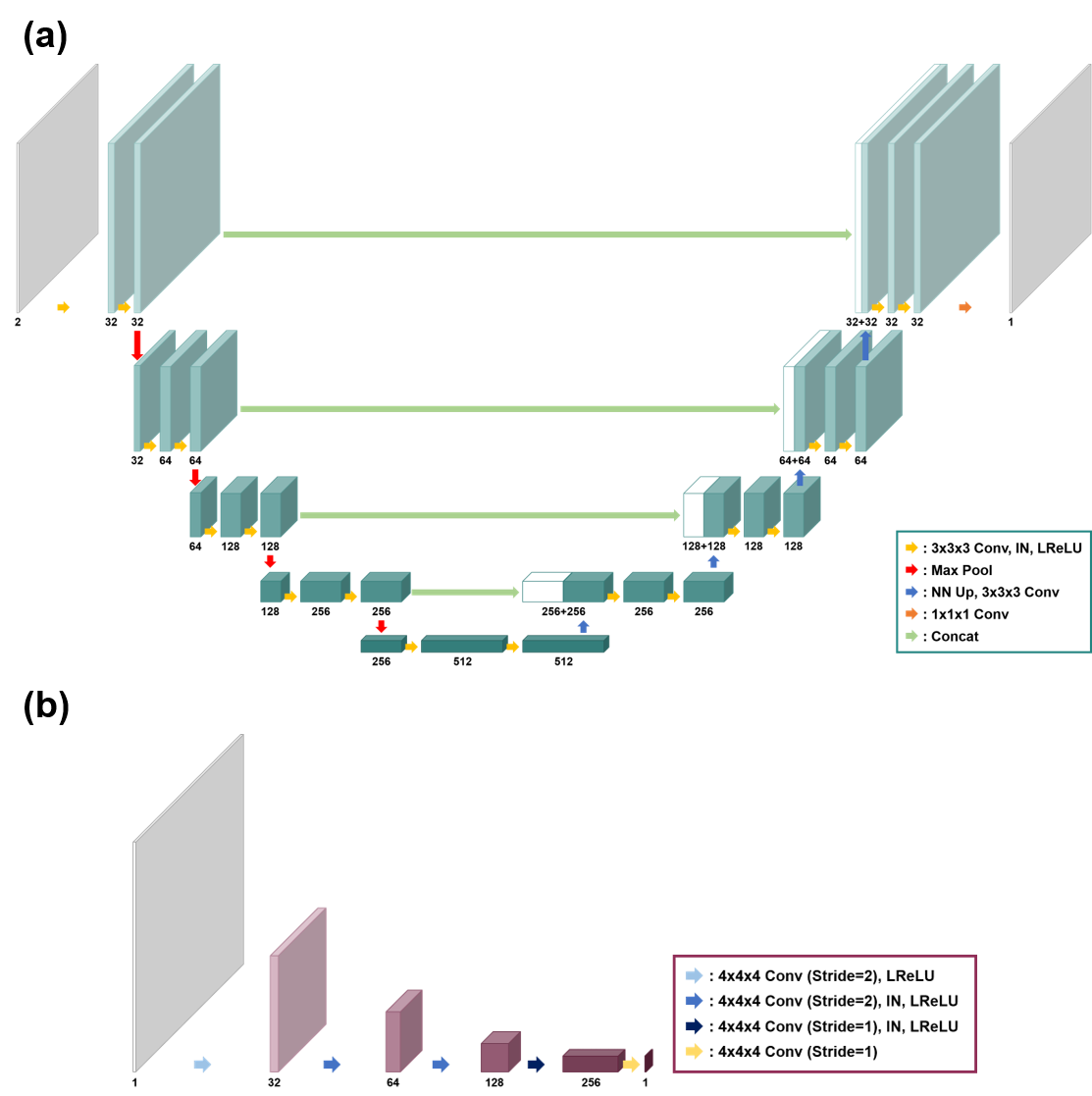}}
    \caption{(a) The generator and (b) the discriminator architecture.}
    \vspace{-0.5cm}
    \label{fig:Networks}
\end{figure}

Fig. \ref{fig:Networks}(a) shows the architecture of our generator.
We use a 3D U-Net \cite{ronneberger2015u} structure to construct our generator.
Because of the concatenation of the magnitude and phase images, our generator has two input channels.
The generator contains 3$\times$3$\times$3 convolution, instance normalization \cite{ulyanov2016instance}, leaky ReLU, and nearest-neighbor upsampling layers.
In addition, the skip connection via channel concatenation is used for the generator.
At the end of the network, the 1$\times$1$\times$1 convolution produces the final reconstructed QSM.

We use patchGAN discriminator \cite{isola2017image,zhu2017unpaired} as shown in Fig. \ref{fig:Networks}(b).
The discriminator consists of 4$\times$4$\times$4 convolution, instance normalization, and leaky ReLU.
Inputs of the discriminator are real susceptibility maps or generated susceptibility maps that are multiplied with brain masks.

\begin{figure*}[!t]
    \centerline{\includegraphics[width=0.8\linewidth]{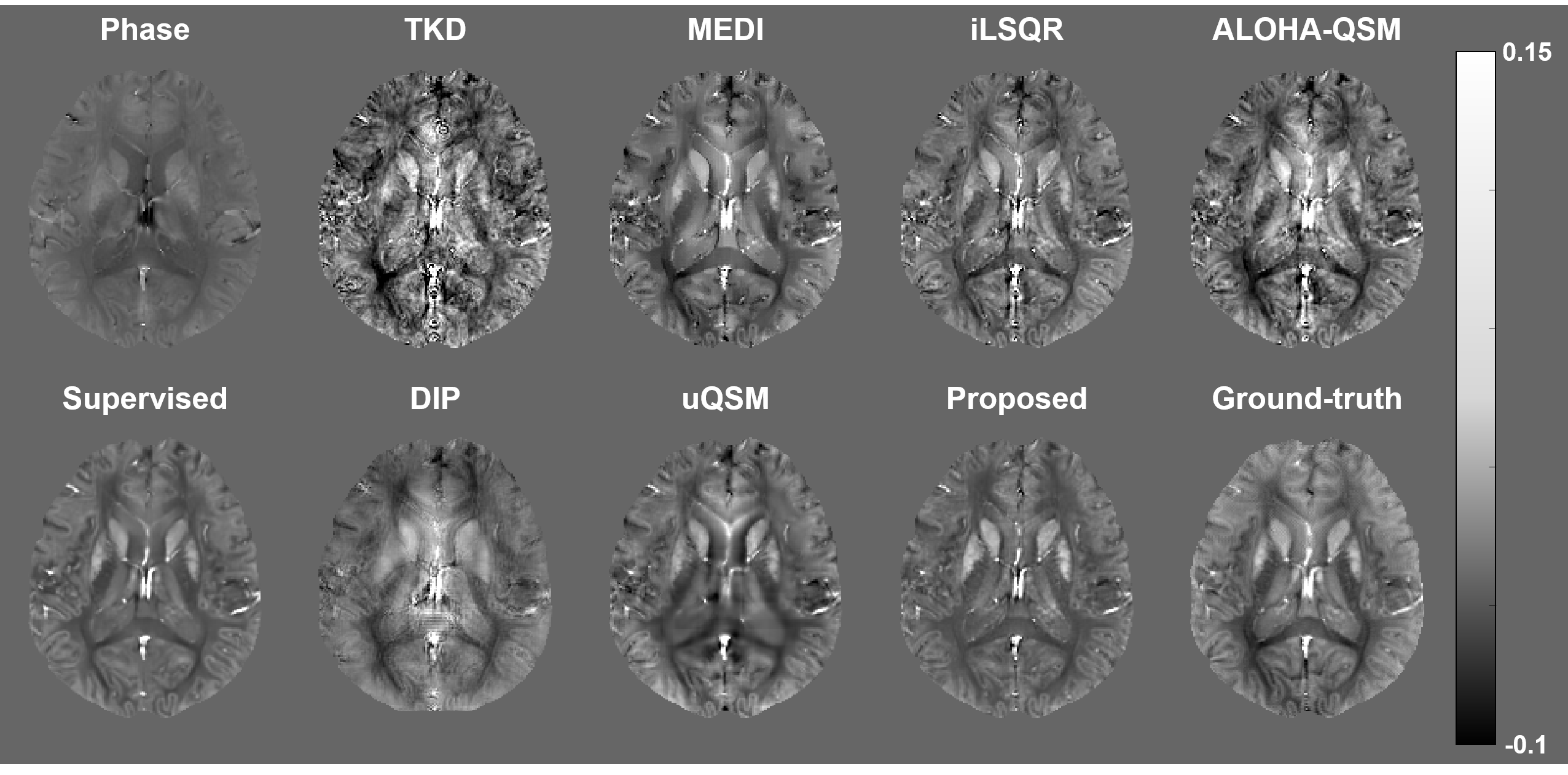}}
    \caption{QSM reconstruction results of Cornell data using various methods.}
    \vspace{-0.5cm}
    \label{fig:results_paired}
\end{figure*}

\subsection{Experimental Details}
Since there are only 8 phase images and 3 QSM label data for network training, we use random patches during training to increase the amount of data.
During one epoch, we extract a total of 3000 phase and unpaired QSM random patches of size 64$\times$64$\times$64, respectively.
In the inference step, a phase volume is cropped into patches of size 64$\times$64$\times$64 with a stride of 16$\times$16$\times$16, and all patch inference results are combined to reconstruct the entire QSM volume.
Moreover, we apply data augmentation by flipping with respect to each axis, and rotating in a plane perpendicular to the direction of the main magnetic field.

In our experiments, we use Adam optimizer with $\beta_1=0.5$, $\beta_2=0.999$, and learning rate of 0.00001.
Also, we choose $\gamma=10$, $\eta=1$, and $\rho=0.1$ for \eqref{eq:cost_final}.
Our cycleGAN is trained for 50 epochs, and implemented in Python by TensorFlow.

\subsection{Comparison Methods}
We compare our algorithm with several conventional methods to verify the performance of our algorithm.
First, TKD \cite{shmueli2009magnetic,schweser2013toward} is used to replace the values in the conical surfaces of the dipole kernel with a certain threshold value:
\begin{eqnarray}
\hat d(\Vec{k};a)=
\begin{cases}
d(\Vec{k}) & \hat |d(\Vec{k})|>a\\
a\cdot\mbox{sign}(d(\Vec{k})) & \hat |d(\Vec{k})|\leq a
\end{cases}
\end{eqnarray}
where $a$ is a threshold value.
Next, MEDI \cite{liu2011morphology,liu2012accuracy} is also compared with our algorithm.
The QSM reconstruction by MEDI can be formulated as follows:
\begin{eqnarray}
\min_\chi\|W(b-\Fc^{-1}\hat d\Fc\chi)\|^2_2+\lambda\|M\nabla\chi\|_1
\end{eqnarray}
where $W$ is the structural weight matrix which is derived by the magnitude image, and $M$ is the binary mask that contains the edge information of the magnitude image.
In addition, we compare cycleQSM with iLSQR \cite{li2015method} and ALOHA-QSM \cite{ahn2020quantitative}.
In our comparison experiments, we use $a=0.1$ for TKD, $\lambda=600$ for MEDI, and 30 iteration steps for iLSQR, respectively.
Also, hyper-parameters for ALOHA-QSM are set to $\lambda=10^{1.5}$, $\mu=10^{-1.5}$ for Cornell data, and $\lambda=10^{2.4}$, $\mu=10^{-2.2}$ for ALOHA-QSM data, respectively.

We also compare our method with other deep learning methods.
Supervised learning using the same U-Net network  is compared with our method.
The network in supervised learning is trained using $L_1$ loss during 50 epochs with the learning rate of 0.0001.
Next, deep image prior (DIP) \cite{ulyanov2018deep} is used for dipole inversion.
DIP is optimized for each volume without training, and the optimization in DIP can be formulated as follows:
\begin{eqnarray}\label{eq:DIP}
\min_\chi\|W(e^{j\Fc^{-1}\hat d\Fc\chi}-e^{jb})\|_1+\lambda\|\nabla\chi\|_1 \ ,
\end{eqnarray}
where $W$ is a noise weighting factor that is obtained from the magnitude image.
We use $\lambda$ in \eqref{eq:DIP} as 0.001 for the best performance.
The network architecture in DIP is same as our generator, but we reduce the number of channels by half due to the GPU memory limitation.
In addition, uQSM \cite{liu2020deep} is used for comparison as another unsupervised learning method.
We use the network architecture in \cite{liu2020deep} for uQSM. 
uQSM minimizes the loss in \eqref{eq:DIP} with $\lambda=0.001$, but it is trained with training data first, then the trained network is used for the reconstruction of test data.
uQSM is trained during 50 epochs with the learning rate of 0.0001.

To evaluate algorithms, we use the peak signal-to-noise ratio (PSNR) and the structural similarity index metrics (SSIM) as quantitative metrics.
PSNR and SSIM are measured for 3D volume of Cornell data which has ground-truth QSM.
Also, we use the root mean square error (RMSE) which is calculated as follows:
\begin{eqnarray}
RMSE = \sqrt{\frac{\sum_{i=1}^N(\chi_i-\tilde\chi_i)^2}{N}} \ ,
\end{eqnarray}
where $N$ is the number of pixels which in the brain mask, and $\chi_i$ ans $\tilde\chi_i$ are pixel intensities of ground-truth and reconstructed QSM.

\begin{figure*}[!t]
    \centerline{\includegraphics[width=0.9\linewidth]{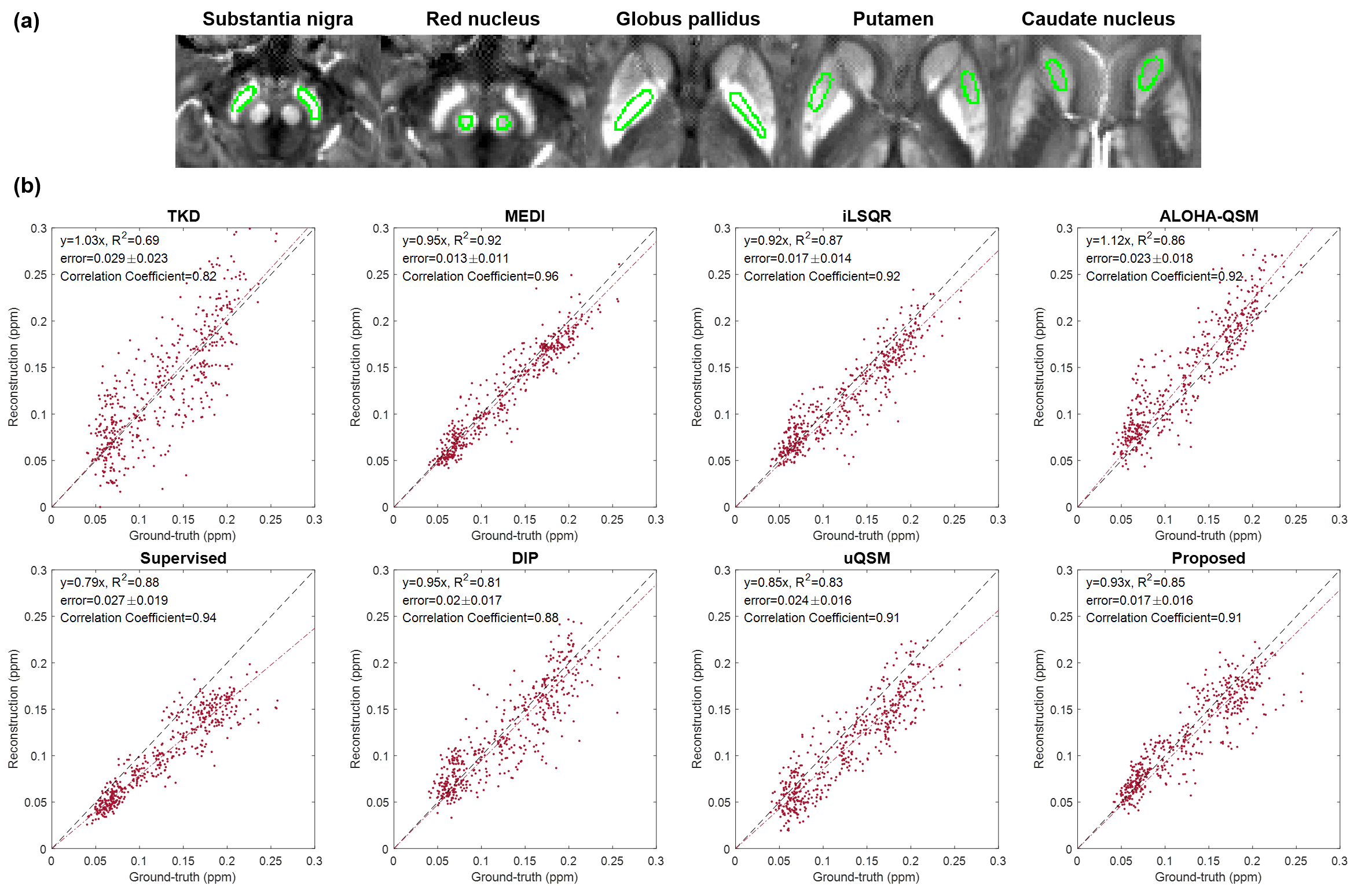}}
    \caption{(a) Deep gray matter structures. The susceptibility values in the green area are extracted for scatter plots and statistical analysis.
    (b) Linear regression between the ground-truth susceptibility values and reconstructed susceptibility values.
    The black dashed line shows the graph of $y=x$, and the red dash-dotted line shows the graph of linear regression results.
    The equation of the red dash-dotted line, $R^2$ value, fitting error, and correlation coefficient are written in the upper left corner of each graph.
    The error (mean$\pm$standard deviation) in the upper left corner is calculated as follows: $|\chi_{COSMOS}-\chi_{Reconstruction}|$.}
    \vspace{-0.5cm}
    \label{fig:results_graph}
\end{figure*}

\section{Experimental Results}\label{sec:result}
First, we compare various QSM reconstruction methods for Cornell data in Table \ref{tbl:dataset} which contains ground-truth QSM data.
Fig. \ref{fig:results_paired} shows the phase image, the ground-truth QSM, and reconstructed QSM by using conventional and deep learning based QSM reconstruction methods.
As shown in Fig. \ref{fig:results_paired}, the reconstructed QSM from TKD has severe streaking artifacts.
MEDI reconstructs susceptibility maps without the streaking artifact, but the output of MEDI is extremely smoothed.
iLSQR and ALOHA-QSM reconstruct more realistic QSM, but some streaking artifacts still remain in the reconstructed QSMs.
Next, supervised learning shows similar output to ground-truth without streaking artifacts.
The output of DIP is smoothed and some susceptibility values are not restored.
Also, DIP requires very long reconstruction time compared to other methods because it has to be optimized to each volume.
uQSM can reconstruct QSM without label data, but uQSM generate smoothed output so that some of structures in the reconstructed QSM are not recognizable.
On the other hand, our method reconstructs QSM that are close to the ground-truth data.
In addition, the proposed method requires similar reconstruction time as supervised learning.

The quantitative metric values of various QSM reconstruction methods are shown in Table \ref{tbl:metric}.
As shown in Table \ref{tbl:metric}, conventional methods show relatively low quantitative metric values due to artifacts in the reconstructed QSM or over-smoothed output.
Supervised learning shows the highest quantitative results because it is trained to minimize $L_1$ loss between the output and label images.
DIP and uQSM shows similar metric values to conventional methods because they show blurry output.
Meanwhile, our method shows comparable quantitative results to supervised learning.
From quantitative evaluation, we confirm that our method outperforms the conventional methods and shows competitive results compared to supervised learning.

\begin{table}[!ht]
    \centering
	\caption{Quantitative evaluation of various QSM reconstruction methods.
	The values in the table are calculated within the whole volume.}
	\label{tbl:metric}
	\resizebox{0.48\textwidth}{!}{
		\begin{tabular}{c | c | c  c  c} 
			\toprule
			\multicolumn{2}{c|}{ } 				                                                            & PSNR (dB) & SSIM	    & RMSE (\%) \\ \midrule\midrule
			\multirow{4}*{Conventional methods}	    & TKD \cite{shmueli2009magnetic,schweser2013toward}     & 38.9398   & 0.9625    & 4.7970    \\
			\ 						                & MEDI \cite{liu2011morphology,liu2012accuracy}         & 42.8732   & 0.9840    & 3.0674    \\
			\                                       & iLSQR \cite{li2015method}                             & 42.8234   & 0.9833    & 3.0850    \\
			\                                       & ALOHA-QSM \cite{ahn2020quantitative}                  & 41.2693   & 0.9778    & 3.6895    \\ \midrule
			\multirow{4}*{Deep learning methods}	& Supervised                                            & 43.4971	& 0.9899    & 2.8548	\\
			\ 						                & DIP \cite{ulyanov2018deep}		                    & 41.3845	& 0.9817	& 3.6409	\\
			\	                                    & uQSM \cite{liu2020deep}	                            & 42.7811	& 0.9866	& 3.1001	\\
			\						                & Proposed		                                        & 43.4656	& 0.9890    & 2.8652	\\ \bottomrule
		\end{tabular}
	}
\end{table}

\begin{table}[!b]
    \centering
	\caption{The slope, $R^2$ value, mean error, and correlation coefficient (Corr) of various QSM reconstruction methods with Cornell data.
	The susceptibility values in gray matter structures are extracted for statistical analysis.
    The error (mean$\pm$standard deviation) is calculated as follows: $|\chi_{COSMOS}-\chi_{Reconstruction}|$.}
	\label{tbl:linear}
	\resizebox{0.48\textwidth}{!}{
		\begin{tabular}{c | c | c  c  c  c} 
			\toprule
			\multicolumn{2}{c|}{ } 				                                                            & Slope     & $R^2$	    & Error   & Corr    \\ \midrule\midrule
			\multirow{4}*{Conventional methods}	    & TKD \cite{shmueli2009magnetic,schweser2013toward}     & 1.03      & 0.69      & 0.029   & 0.82    \\
			\ 						                & MEDI \cite{liu2011morphology,liu2012accuracy}         & 0.95      & 0.92      & 0.013   & 0.96    \\
			\                                       & iLSQR \cite{li2015method}                             & 0.92      & 0.87      & 0.017   & 0.92    \\
			\                                       & ALOHA-QSM \cite{ahn2020quantitative}                  & 1.12      & 0.86      & 0.023   & 0.92    \\ \midrule
			\multirow{4}*{Deep learning methods}	& Supervised                                            & 0.79	    & 0.88      & 0.027	  & 0.94    \\
			\ 						                & DIP \cite{ulyanov2018deep}		                    & 0.95	    & 0.81	    & 0.020	  & 0.88    \\
			\	                                    & uQSM \cite{liu2020deep}	                            & 0.85	    & 0.83	    & 0.024	  & 0.91    \\
			\						                & Proposed		                                        & 0.93	    & 0.85      & 0.017	  & 0.91    \\ \bottomrule
		\end{tabular}
	}
	\vspace{-0.5cm}
\end{table}

\begin{figure*}[!hbt]
    \centerline{\includegraphics[width=0.99\linewidth]{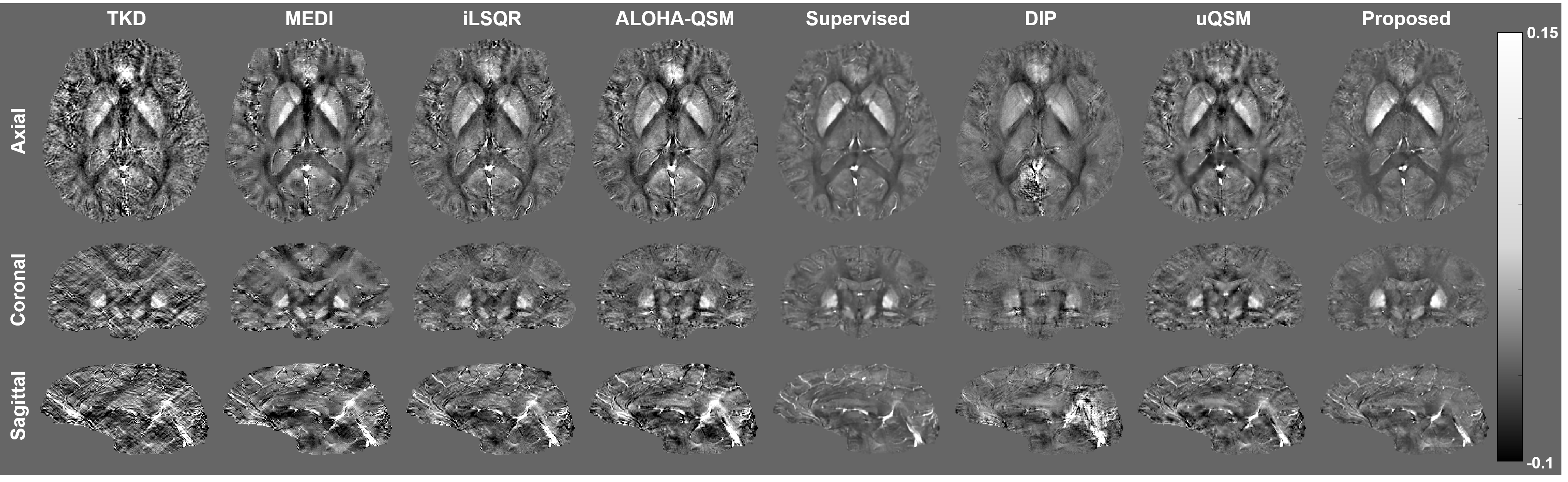}}
    \caption{QSM reconstruction results of ALOHA-QSM data using various methods.}
    \vspace{-0.5cm}
    \label{fig:results_unpaired1}
\end{figure*}

To verify that our method restores accurate susceptibility values in gray matter structures, we extract susceptibility values in gray matter structures in Fig. \ref{fig:results_graph}(a), and compare them with ground-truth susceptibility values through linear regression and correlation coefficient.
Fig. \ref{fig:results_graph}(b) shows the scatter plots for the ground-truth and the reconstruction QSM value pairs, and Table \ref{tbl:linear} summarizes the statistical analysis results between susceptibility values of ground-truth QSM and reconstructed QSM.
Because the reconstructed QSM by TKD has severe streaking artifacts, there are many outliers in the regression graph of TKD.
Therefore, the $R^2$ value and correlation coefficient in TKD is low and the error of TKD is higher than other methods.
Next, MEDI shows the best performance in terms of linear regression and correlation coefficient.
The linear regression result of MEDI is close to $y=x$ and has large $R^2$ values.
Also, the correlation coefficient of MEDI is the highest among QSM reconstruction methods.
However, note that this accuracy is compromised by the overly smoothed reconstruction results in Fig. \ref{fig:results_paired}.
The iLSQR shows slightly lower performance than MEDI, but the error of iLSQR is relatively small compared to other methods.
The slope of ALOHA-QSM is higher than 1, which indicates there are some overestimated susceptibility values when using ALOHA-QSM method.

On the other hand, many points in the graph of supervised learning are below the line of $y=x$.
Furthermore, the slope of supervised learning is lower than 1 and it shows large error due to the underestimation, which confirms the reported findings \cite{jung2020overview}.
Since DIP cannot reconstruct accurate QSM, there exist many underestimated or overestimated points in the graph.
Also, DIP has low $R^2$ value and correlation coefficient compared to other QSM reconstruction methods.
uQSM also underestimate many susceptibility values, so it shows large error value.
Meanwhile, our method reconstructs accurate susceptibility values, thus it has fewer underestimated points than supervised learning.
Furthermore, the slope of our method is close to 1, and the error is lower than other methods except MEDI.
From these results, we reveal that our method provides comparable quantitative results with the classical methods and outperforms other deep learning methods.

We also employ QSM reconstruction algorithms to ALOHA-QSM data which does not contain ground-truth QSM.
Fig. \ref{fig:results_unpaired1} shows the QSM reconstruction results of ALOHA-QSM data using various QSM reconstruction methods.
As shown in Fig. \ref{fig:results_unpaired1}, TKD generates severe streaking artifacts, so the quality of reconstructed QSM is degraded and it is difficult to distinguish anatomical structures in the brain.
MEDI also shows smoothed reconstruction results, and checkerboard artifacts are remarkable when MEDI is applied for ALOHA-QSM data.
iLSQR and ALOHA-QSM show more realistic reconstruction than TKD or MEDI, but slight noises were seen in the reconstructed susceptibility maps.
Supervised learning does not show artifacts, but the results of supervised learning seem blurry.
Next, like the previous experiment using Cornell data, DIP shows inferior results.
Especially, the severe artifacts are appeared in the sagittal plain.
uQSM generates artifacts and blurred region in reconstructed QSM.
On the other hand, the proposed method does not show the artifact or noise in reconstructed QSM.
Moreover, the contrast and brain structures are recovered accurately when using our method.

\begin{figure}[!hbt]
    \centerline{\includegraphics[width=0.9\linewidth]{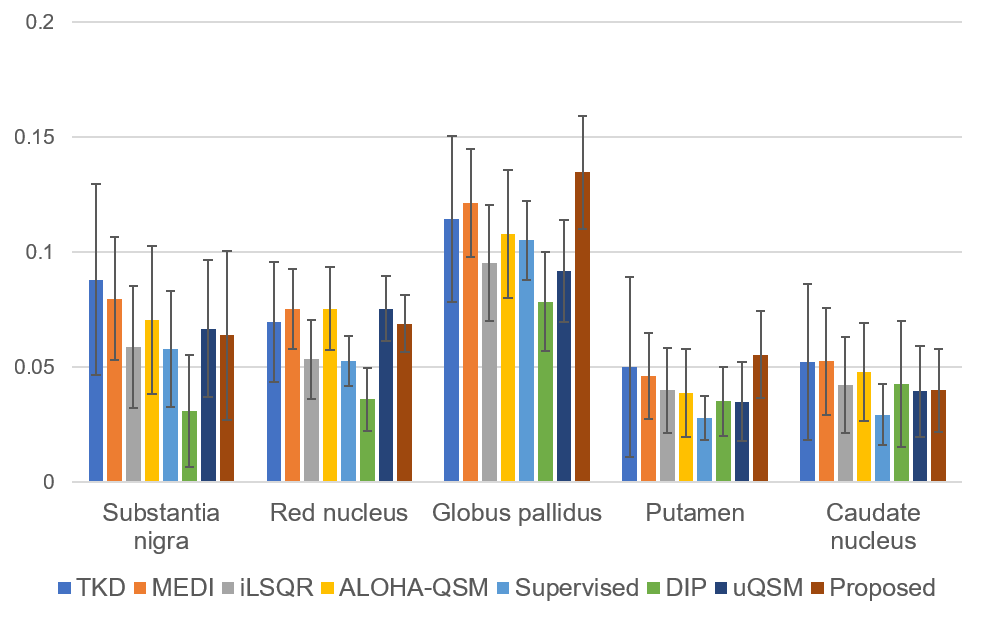}}
    \caption{Mean susceptibility value of each deep gray matter structure in the reconstructed QSMs using various methods with ALOHA-QSM data.
    The error bars indicate the standard deviation of susceptibility values.}
    \vspace{-0.5cm}
    \label{fig:results_histogram_unpaired}
\end{figure}

Because ALOHA-QSM data does not contain ground-truth QSM data, we compare mean susceptibility values of each brain structure in reconstructed susceptibility maps to verify that our method provides accurate susceptibility values.
Fig. \ref{fig:results_histogram_unpaired} shows the mean susceptibility values of brain structures in reconstructed QSM.
Deep learning based algorithms show lower susceptibility values than conventional algorithms on average.
Especially, supervised learning shows the lowest mean susceptibility value in the putamen and caudate nucleus.
On the contrary, the proposed method shows similar mean susceptibility values with conventional methods.
Therefore, we believe that our method provides proper susceptibility values.

\section{Discussion}\label{sec:discussion}
To verify our cycleGAN structure, we compare with other cycleGAN structures in \cite{sim2019optimal}.
The first structure is the conventional cycleGAN with two deep generators and two discriminators.
Two generators and discriminators in the conventional cycleGAN have same architecture as ours.
Next, we also compare our method to cycleGAN with linear {\em unknown} kernel as proposed in \cite{sim2019optimal} and \cite{lim2020cyclegan}.
Because the relationship between the phase image and QSM can be formulated as the convolution operator in the spatial domain, it is possible to replace the generator for producing the phase image from the QSM to one linear convolution kernel that can be trained.
We set the size of the convolution kernel for generating the phase image from the QSM as 20$\times$20$\times$20.

\begin{figure}[!t]
    \centerline{\includegraphics[width=0.99\linewidth]{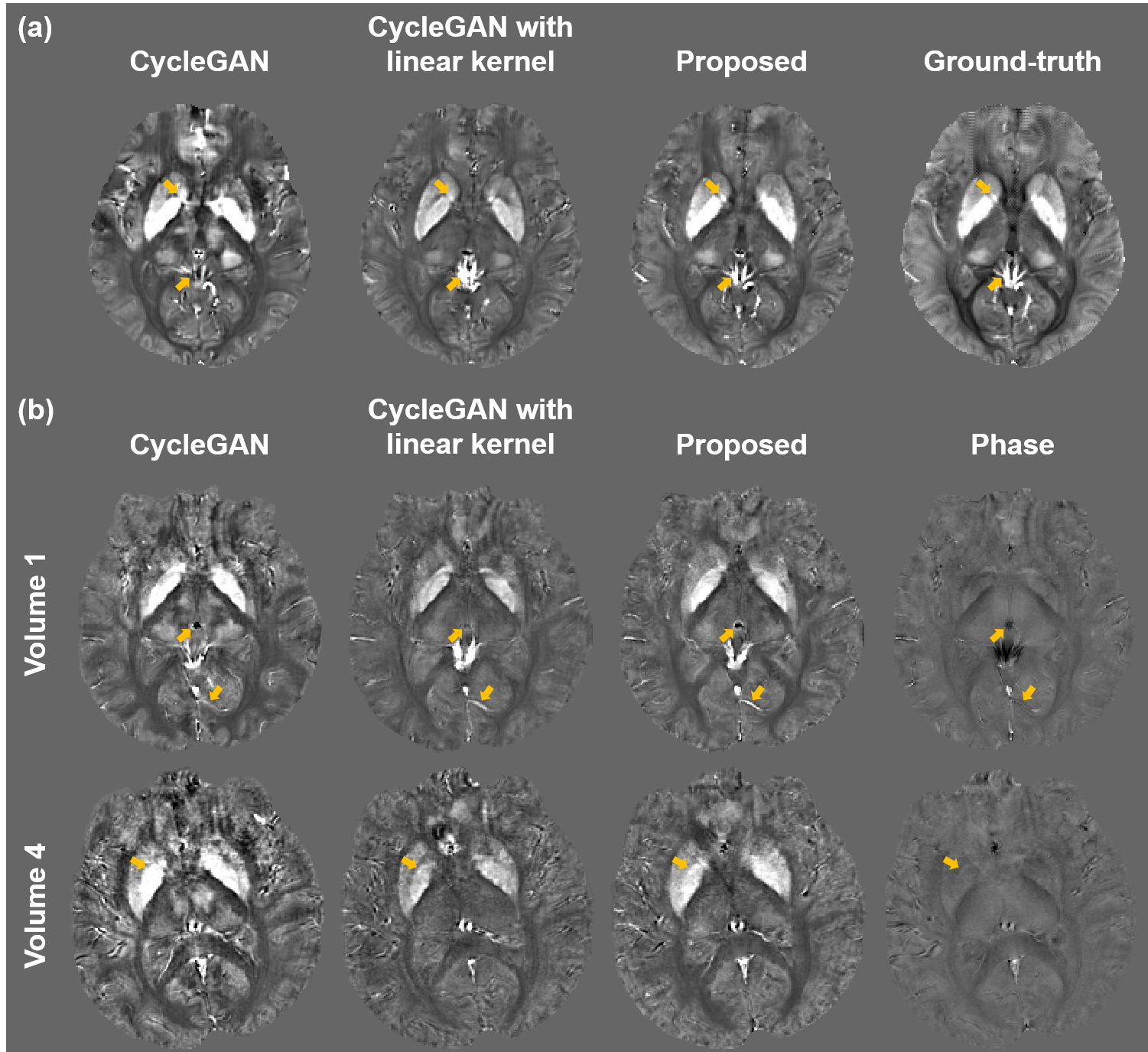}}
    \caption{QSM reconstruction results of (a) Cornell data and (b) ALOHA-QSM data using various cycleGAN structures.
    Yellow arrows indicate artifacts or errors in the results of cycleGAN and cycleGAN with linear kernel.}
    \vspace{-0.5cm}
    \label{fig:results_cycle}
\end{figure}

Fig. \ref{fig:results_cycle}(a) and (b) show the reconstruction results using various cycleGAN structures.
As shown in Fig. \ref{fig:results_cycle}(a), the reconstruction result by the conventional cycleGAN has different contrast with the ground-truth QSM.
Also, some susceptibility values are overestimated. 
Since conventional cycleGAN has four neural networks which have to be optimized simultaneously, it requires large training data for stable convergence. 
Therefore, the difficulty of convergence of conventional cycleGAN leads to the inferior reconstruction results.
Next, cycleGAN with linear kernel is unable to restore appropriate susceptibility values. 
Because each data set that were used for training has different resolution, the dipole kernels of each data set are also different.
Therefore, it is difficult to replace all dipole kernels to single linear convolution, and it makes inappropriate reconstruction in cycleGAN with linear kernel.
On the other hand, the proposed cycleGAN reconstructes precise susceptibility values without any artifacts.
Because our cycleGAN employs deterministic dipole kernel that is generated depending on the resolution of each data, it is possible to train the networks and reconstruct accurate QSM with only one generator and one discriminator by utilizing all training data at different spatial resolution.
In Fig. \ref{fig:results_cycle}(b), various cycleGAN structures are also applied for ALOHA-QSM data. 
Compared to other cycleGAN structures, our cycleGAN reconstructs accurate QSM without the artifact and noise.

We also compare quantitative metric values of each cycleGAN structure using Cornell data set.
From Table \ref{tbl:cycle}, we demonstrate that quantitative metric values of conventional cycleGAN and cycleGAN with linear kernel are lower than those of our method.

\begin{table}[!ht]
    \centering
	\caption{Quantitative evaluation of various cycleGAN structures.
	The values in the table are calculated within the whole volume.}
	\label{tbl:cycle}
	\resizebox{0.45\textwidth}{!}{
		\begin{tabular}{c | c  c  c} 
			\toprule
			\ 				                    & PSNR (dB) & SSIM	    & RMSE (\%) \\ \midrule\midrule
			\ CycleGAN                          & 40.1006	& 0.9753    & 4.2208	\\
		    \ CycleGAN with linear kernel       & 41.9220	& 0.9825	& 3.4224	\\
		    \ Proposed		                    & 43.4656	& 0.9890    & 2.8652	\\ \bottomrule
		\end{tabular}
	}
	\vspace{-0.5cm}
\end{table}


\section{Conclusion}\label{sec:conclusion}
In this paper, we developed a novel unsupervised QSM reconstruction method called cycleQSM using physics-informed cycleGAN which is derived from optimal transport perspective.
Due to the known dipole kernel, our cycleQSM has only a single pair of generator and discriminator, which makes the training much stable.

In our experiments, conventional QSM reconstruction methods showed inferior qualitative results compared to the proposed method due to the streaking artifacts or smoothing.
Furthermore, the reconstruction time of conventional methods is relatively long compared to deep learning methods due to the high computational complexity.
Supervised learning 
generated blurry output and many susceptibility values were underestimated.
%

Although the proposed method was trained without the matched pair of phase and QSM,
we demonstrated that the proposed method shows comparable quantitative results compared to supervised learning.
Furthermore, we confirmed that our method provides accurate susceptibility values without underestimation.
Therefore, we believe that cycleQSM can be a powerful framework for QSM reconstruction. 


\bibliographystyle{IEEEtran}
\bibliography{ref}

\begin{thebibliography}{10}
\providecommand{\url}[1]{#1}
\csname url@samestyle\endcsname
\providecommand{\newblock}{\relax}
\providecommand{\bibinfo}[2]{#2}
\providecommand{\BIBentrySTDinterwordspacing}{\spaceskip=0pt\relax}
\providecommand{\BIBentryALTinterwordstretchfactor}{4}
\providecommand{\BIBentryALTinterwordspacing}{\spaceskip=\fontdimen2\font plus
\BIBentryALTinterwordstretchfactor\fontdimen3\font minus
  \fontdimen4\font\relax}
\providecommand{\BIBforeignlanguage}[2]{{%
\expandafter\ifx\csname l@#1\endcsname\relax
\typeout{** WARNING: IEEEtran.bst: No hyphenation pattern has been}%
\typeout{** loaded for the language `#1'. Using the pattern for}%
\typeout{** the default language instead.}%
\else
\language=\csname l@#1\endcsname
\fi
#2}}
\providecommand{\BIBdecl}{\relax}
\BIBdecl

\bibitem{haacke2015quantitative}
E.~M. Haacke, S.~Liu, S.~Buch, W.~Zheng, D.~Wu, and Y.~Ye, ``Quantitative
  susceptibility mapping: current status and future directions,''
  \emph{Magnetic resonance imaging}, vol.~33, no.~1, pp. 1--25, 2015.

\bibitem{wang2015quantitative}
Y.~Wang and T.~Liu, ``Quantitative susceptibility mapping ({QSM}): decoding
  {MRI} data for a tissue magnetic biomarker,'' \emph{Magnetic resonance in
  medicine}, vol.~73, no.~1, pp. 82--101, 2015.

\bibitem{bilgic2012mri}
B.~Bilgic, A.~Pfefferbaum, T.~Rohlfing, E.~V. Sullivan, and E.~Adalsteinsson,
  ``{MRI} estimates of brain iron concentration in normal aging using
  quantitative susceptibility mapping,'' \emph{Neuroimage}, vol.~59, no.~3, pp.
  2625--2635, 2012.

\bibitem{deistung2013quantitative}
A.~Deistung, F.~Schweser, B.~Wiestler, M.~Abello, M.~Roethke, F.~Sahm, W.~Wick,
  A.~M. Nagel, S.~Heiland, H.-P. Schlemmer \emph{et~al.}, ``Quantitative
  susceptibility mapping differentiates between blood depositions and
  calcifications in patients with glioblastoma,'' \emph{PloS one}, vol.~8,
  no.~3, p. e57924, 2013.

\bibitem{langkammer2013quantitative}
C.~Langkammer, T.~Liu, M.~Khalil, C.~Enzinger, M.~Jehna, S.~Fuchs, F.~Fazekas,
  Y.~Wang, and S.~Ropele, ``Quantitative susceptibility mapping in multiple
  sclerosis,'' \emph{Radiology}, vol. 267, no.~2, pp. 551--559, 2013.

\bibitem{barbosa2015quantifying}
J.~H.~O. Barbosa, A.~C. Santos, V.~Tumas, M.~Liu, W.~Zheng, E.~M. Haacke, and
  C.~E.~G. Salmon, ``Quantifying brain iron deposition in patients with
  parkinson's disease using quantitative susceptibility mapping, r2 and r2,''
  \emph{Magnetic resonance imaging}, vol.~33, no.~5, pp. 559--565, 2015.

\bibitem{liu2009calculation}
T.~Liu, P.~Spincemaille, L.~De~Rochefort, B.~Kressler, and Y.~Wang,
  ``{Calculation of susceptibility through multiple orientation sampling
  (COSMOS): a method for conditioning the inverse problem from measured
  magnetic field map to susceptibility source image in MRI},'' \emph{Magnetic
  Resonance in Medicine: An Official Journal of the International Society for
  Magnetic Resonance in Medicine}, vol.~61, no.~1, pp. 196--204, 2009.

\bibitem{shmueli2009magnetic}
K.~Shmueli, J.~A. de~Zwart, P.~van Gelderen, T.-Q. Li, S.~J. Dodd, and J.~H.
  Duyn, ``Magnetic susceptibility mapping of brain tissue in vivo using {MRI}
  phase data,'' \emph{Magnetic Resonance in Medicine: An Official Journal of
  the International Society for Magnetic Resonance in Medicine}, vol.~62,
  no.~6, pp. 1510--1522, 2009.

\bibitem{schweser2013toward}
F.~Schweser, A.~Deistung, K.~Sommer, and J.~R. Reichenbach, ``Toward online
  reconstruction of quantitative susceptibility maps: superfast dipole
  inversion,'' \emph{Magnetic resonance in medicine}, vol.~69, no.~6, pp.
  1581--1593, 2013.

\bibitem{liu2011morphology}
T.~Liu, J.~Liu, L.~De~Rochefort, P.~Spincemaille, I.~Khalidov, J.~R. Ledoux,
  and Y.~Wang, ``{Morphology enabled dipole inversion (MEDI) from a
  single-angle acquisition: comparison with COSMOS in human brain imaging},''
  \emph{Magnetic resonance in medicine}, vol.~66, no.~3, pp. 777--783, 2011.

\bibitem{liu2012accuracy}
T.~Liu, W.~Xu, P.~Spincemaille, A.~S. Avestimehr, and Y.~Wang, ``{Accuracy of
  the morphology enabled dipole inversion ({MEDI}) algorithm for quantitative
  susceptibility mapping in MRI},'' \emph{IEEE transactions on medical
  imaging}, vol.~31, no.~3, pp. 816--824, 2012.

\bibitem{wu2012whole}
B.~Wu, W.~Li, A.~Guidon, and C.~Liu, ``Whole brain susceptibility mapping using
  compressed sensing,'' \emph{Magnetic resonance in medicine}, vol.~67, no.~1,
  pp. 137--147, 2012.

\bibitem{ahn2020quantitative}
H.-S. Ahn, S.-H. Park, and J.~C. Ye, ``Quantitative susceptibility map
  reconstruction using annihilating filter-based low-rank hankel matrix
  approach,'' \emph{Magnetic Resonance in Medicine}, vol.~83, no.~3, pp.
  858--871, 2020.

\bibitem{kang2017deep}
E.~Kang, J.~Min, and J.~C. Ye, ``A deep convolutional neural network using
  directional wavelets for low-dose x-ray ct reconstruction,'' \emph{Medical
  physics}, vol.~44, no.~10, pp. e360--e375, 2017.

\bibitem{gong2018deep}
E.~Gong, J.~M. Pauly, M.~Wintermark, and G.~Zaharchuk, ``Deep learning enables
  reduced gadolinium dose for contrast-enhanced brain {MRI},'' \emph{Journal of
  magnetic resonance imaging}, vol.~48, no.~2, pp. 330--340, 2018.

\bibitem{cha2020geometric}
E.~Cha, G.~Oh, and J.~C. Ye, ``Geometric approaches to increase the
  expressivity of deep neural networks for mr reconstruction,'' \emph{IEEE
  Journal of Selected Topics in Signal Processing}, 2020.

\bibitem{khan2020adaptive}
S.~Khan, J.~Huh, and J.~C. Ye, ``Adaptive and compressive beamforming using
  deep learning for medical ultrasound,'' \emph{IEEE Transactions on
  Ultrasonics, Ferroelectrics, and Frequency Control}, 2020.

\bibitem{yoon2018quantitative}
J.~Yoon, E.~Gong, I.~Chatnuntawech, B.~Bilgic, J.~Lee, W.~Jung, J.~Ko, H.~Jung,
  K.~Setsompop, G.~Zaharchuk \emph{et~al.}, ``Quantitative susceptibility
  mapping using deep neural network: {QSM}net,'' \emph{Neuroimage}, vol. 179,
  pp. 199--206, 2018.

\bibitem{bollmann2019deepqsm}
S.~Bollmann, K.~G.~B. Rasmussen, M.~Kristensen, R.~G. Blendal, L.~R.
  {\O}stergaard, M.~Plocharski, K.~O'Brien, C.~Langkammer, A.~Janke, and
  M.~Barth, ``Deep{QSM}-using deep learning to solve the dipole inversion for
  quantitative susceptibility mapping,'' \emph{Neuroimage}, vol. 195, pp.
  373--383, 2019.

\bibitem{chen2020qsmgan}
Y.~Chen, A.~Jakary, S.~Avadiappan, C.~P. Hess, and J.~M. Lupo, ``{QSMGAN}:
  improved quantitative susceptibility mapping using 3d generative adversarial
  networks with increased receptive field,'' \emph{NeuroImage}, vol. 207, p.
  116389, 2020.

\bibitem{gao2020xqsm}
Y.~Gao, X.~Zhu, S.~Crozier, F.~Liu, and H.~Sun, ``x{QSM}-quantitative
  susceptibility mapping with octave convolutional neural networks,''
  \emph{arXiv preprint arXiv:2004.06281}, 2020.

\bibitem{liu2020deep}
J.~Liu, ``Deep learning for quantitative susceptibility mapping without
  labels,'' \emph{arXiv preprint arXiv:2004.06259}, 2020.

\bibitem{liu2020weakly}
J.~Liu and K.~M. Koch, ``Weakly-supervised learning for single-step
  quantitative susceptibility mapping,'' \emph{arXiv preprint
  arXiv:2008.06187}, 2020.

\bibitem{polak2020nonlinear}
D.~Polak, I.~Chatnuntawech, J.~Yoon, S.~S. Iyer, C.~Milovic, J.~Lee,
  P.~Bachert, E.~Adalsteinsson, K.~Setsompop, and B.~Bilgic, ``Nonlinear dipole
  inversion (ndi) enables robust quantitative susceptibility mapping ({QSM}),''
  \emph{NMR in Biomedicine}, p. e4271, 2020.

\bibitem{jung2020overview}
W.~Jung, S.~Bollmann, and J.~Lee, ``Overview of quantitative susceptibility
  mapping using deep learning: Current status, challenges and opportunities,''
  \emph{NMR in Biomedicine}, p. e4292, 2020.

\bibitem{sim2019optimal}
B.~Sim, G.~Oh, J.~Kim, C.~Jung, and J.~C. Ye, ``Optimal transport driven
  {CycleGAN} for unsupervised learning in inverse problems,'' \emph{SIAM J.
  Imaging Sciences (in press), Also available arXiv preprint arXiv:1909.12116},
  2020.

\bibitem{lim2020cyclegan}
S.~Lim, H.~Park, S.-E. Lee, S.~Chang, B.~Sim, and J.~C. Ye, ``Cycle{GAN} with a
  blur kernel for deconvolution microscopy: Optimal transport geometry,''
  \emph{IEEE Transactions on Computational Imaging}, vol.~6, pp. 1127--1138,
  2020.

\bibitem{oh2020unpaired}
G.~Oh, B.~Sim, H.~Chung, L.~Sunwoo, and J.~C. Ye, ``Unpaired deep learning for
  accelerated {MRI} using optimal transport driven {Cycle}{GAN},'' \emph{IEEE
  Transactions on Computational Imaging}, 2020.

\bibitem{schweser2012quantitative}
F.~Schweser, K.~Sommer, A.~Deistung, and J.~R. Reichenbach, ``Quantitative
  susceptibility mapping for investigating subtle susceptibility variations in
  the human brain,'' \emph{Neuroimage}, vol.~62, no.~3, pp. 2083--2100, 2012.

\bibitem{li2015method}
W.~Li, N.~Wang, F.~Yu, H.~Han, W.~Cao, R.~Romero, B.~Tantiwongkosi, T.~Q.
  Duong, and C.~Liu, ``A method for estimating and removing streaking artifacts
  in quantitative susceptibility mapping,'' \emph{Neuroimage}, vol. 108, pp.
  111--122, 2015.

\bibitem{CaRoTa06}
E.~Candes, J.~Romberg, and T.~Tao, ``Robust uncertainty principles: Exact
  signal reconstruction from highly incomplete frequency information,''
  \emph{IEEE Trans. on Information Theory}, vol.~52, no.~2, pp. 489--509, Feb.
  2006.

\bibitem{donoho2006compressed}
D.~L. Donoho, ``Compressed sensing,'' \emph{IEEE Transactions on information
  theory}, vol.~52, no.~4, pp. 1289--1306, 2006.

\bibitem{ronneberger2015u}
O.~Ronneberger, P.~Fischer, and T.~Brox, ``U-net: Convolutional networks for
  biomedical image segmentation,'' in \emph{International Conference on Medical
  Image Computing and Computer-Assisted Intervention}.\hskip 1em plus 0.5em
  minus 0.4em\relax Springer, 2015, pp. 234--241.

\bibitem{ulyanov2018deep}
D.~Ulyanov, A.~Vedaldi, and V.~Lempitsky, ``Deep image prior,'' in
  \emph{Proceedings of the IEEE Conference on Computer Vision and Pattern
  Recognition}, 2018, pp. 9446--9454.

\bibitem{villani2008optimal}
C.~Villani, \emph{Optimal transport: old and new}.\hskip 1em plus 0.5em minus
  0.4em\relax Springer Science \& Business Media, 2008, vol. 338.

\bibitem{martin2017wasserstein}
S.~Martin~Arjovsky and L.~Bottou, ``Wasserstein generative adversarial
  networks,'' in \emph{Proceedings of the 34 th International Conference on
  Machine Learning, Sydney, Australia}, 2017.

\bibitem{mao2017least}
X.~Mao, Q.~Li, H.~Xie, R.~Y. Lau, Z.~Wang, and S.~Paul~Smolley, ``Least squares
  generative adversarial networks,'' in \emph{Proceedings of the IEEE
  international conference on computer vision}, 2017, pp. 2794--2802.

\bibitem{cha2020unpaired}
E.~Cha, H.~Chung, E.~Y. Kim, and J.~C. Ye, ``Unpaired training of deep learning
  {tMRA} for flexible spatio-temporal resolution,'' \emph{IEEE Transactions on
  Medical Imaging}, 2020.

\bibitem{langkammer2018quantitative}
C.~Langkammer, F.~Schweser, K.~Shmueli, C.~Kames, X.~Li, L.~Guo, C.~Milovic,
  J.~Kim, H.~Wei, K.~Bredies \emph{et~al.}, ``Quantitative susceptibility
  mapping: report from the 2016 reconstruction challenge,'' \emph{Magnetic
  resonance in medicine}, vol.~79, no.~3, pp. 1661--1673, 2018.

\bibitem{marques2020qsm}
J.~P. Marques, J.~Meineke, C.~Milovic, B.~Bilgic, K.-S. Chan, R.~Hedouin,
  W.~vand~der Zwaag, C.~Langkammer, and F.~Schweser, ``{QSM} reconstruction
  challenge 2.0 part 1: A realistic in silico head phantom for {MRI} data
  simulation and evaluation of susceptibility mapping procedures,''
  \emph{bioRxiv}, 2020.

\bibitem{smith2002fast}
S.~M. Smith, ``Fast robust automated brain extraction,'' \emph{Human brain
  mapping}, vol.~17, no.~3, pp. 143--155, 2002.

\bibitem{bernstein1994reconstructions}
M.~A. Bernstein, M.~Grgic, T.~J. Brosnan, and N.~J. Pelc, ``Reconstructions of
  phase contrast, phased array multicoil data,'' \emph{Magnetic resonance in
  medicine}, vol.~32, no.~3, pp. 330--334, 1994.

\bibitem{de2008quantitative}
L.~De~Rochefort, R.~Brown, M.~R. Prince, and Y.~Wang, ``Quantitative mr
  susceptibility mapping using piece-wise constant regularized inversion of the
  magnetic field,'' \emph{Magnetic Resonance in Medicine: An Official Journal
  of the International Society for Magnetic Resonance in Medicine}, vol.~60,
  no.~4, pp. 1003--1009, 2008.

\bibitem{kressler2009nonlinear}
B.~Kressler, L.~De~Rochefort, T.~Liu, P.~Spincemaille, Q.~Jiang, and Y.~Wang,
  ``Nonlinear regularization for per voxel estimation of magnetic
  susceptibility distributions from {MRI} field maps,'' \emph{IEEE transactions
  on medical imaging}, vol.~29, no.~2, pp. 273--281, 2009.

\bibitem{liu2013nonlinear}
T.~Liu, C.~Wisnieff, M.~Lou, W.~Chen, P.~Spincemaille, and Y.~Wang, ``Nonlinear
  formulation of the magnetic field to source relationship for robust
  quantitative susceptibility mapping,'' \emph{Magnetic resonance in medicine},
  vol.~69, no.~2, pp. 467--476, 2013.

\bibitem{schofield2003fast}
M.~A. Schofield and Y.~Zhu, ``Fast phase unwrapping algorithm for
  interferometric applications,'' \emph{Optics letters}, vol.~28, no.~14, pp.
  1194--1196, 2003.

\bibitem{li2011quantitative}
W.~Li, B.~Wu, and C.~Liu, ``Quantitative susceptibility mapping of human brain
  reflects spatial variation in tissue composition,'' \emph{Neuroimage},
  vol.~55, no.~4, pp. 1645--1656, 2011.

\bibitem{li2014integrated}
W.~Li, A.~V. Avram, B.~Wu, X.~Xiao, and C.~Liu, ``Integrated laplacian-based
  phase unwrapping and background phase removal for quantitative susceptibility
  mapping,'' \emph{NMR in Biomedicine}, vol.~27, no.~2, pp. 219--227, 2014.

\bibitem{ulyanov2016instance}
D.~Ulyanov, A.~Vedaldi, and V.~Lempitsky, ``Instance normalization: The missing
  ingredient for fast stylization,'' \emph{arXiv preprint arXiv:1607.08022},
  2016.

\bibitem{isola2017image}
P.~Isola, J.-Y. Zhu, T.~Zhou, and A.~A. Efros, ``Image-to-image translation
  with conditional adversarial networks,'' in \emph{Proceedings of the IEEE
  conference on computer vision and pattern recognition}, 2017, pp. 1125--1134.

\bibitem{zhu2017unpaired}
J.-Y. Zhu, T.~Park, P.~Isola, and A.~A. Efros, ``Unpaired image-to-image
  translation using cycle-consistent adversarial networks,'' in
  \emph{Proceedings of the IEEE international conference on computer vision},
  2017, pp. 2223--2232.

\end{thebibliography}

\end{document}